\documentclass{article}

\usepackage[nonatbib]{nips_2018}

\usepackage[utf8]{inputenc} 
\usepackage[T1]{fontenc}    
\usepackage{hyperref}       
\usepackage{url}            
\usepackage{booktabs}       
\usepackage{amsfonts}       
\usepackage{nicefrac}       
\usepackage{microtype}      
\usepackage{authblk}
\usepackage{chngcntr}       
\usepackage{bm}             
\usepackage[pdftex]{graphicx}
\usepackage{color}
\usepackage{footnote}

\begin{document}

\title{Fast Inference in Capsule Networks Using Accumulated Routing Coefficients}

\maketitle

\begin{abstract}
	We present a method for fast inference in Capsule Networks (CapsNets) by taking advantage of a key insight regarding the routing coefficients that link capsules between adjacent network layers. Since the routing coefficients are responsible for assigning object parts to wholes, and an object whole generally contains similar intra-class and dissimilar inter-class parts, the routing coefficients tend to form a unique signature for each object class. For fast inference, a network is first trained in the usual manner using examples from the training dataset. Afterward, the routing coefficients associated with the training examples are accumulated offline and used to create a set of ``master'' routing coefficients. During inference, these master routing coefficients are used in place of the dynamically calculated routing coefficients. Our method effectively replaces the for-loop iterations in the dynamic routing procedure \cite{Sabour_2017} with a single matrix multiply operation, providing a significant boost in inference speed. Compared with the dynamic routing procedure, fast inference decreases the test accuracy for the MNIST, Background MNIST, Fashion MNIST, and Rotated MNIST datasets by less than $0.5\%$ and by approximately $5\%$ for CIFAR10.
\end{abstract}

\section{Introduction}
\label{Section: Introduction}
For object recognition to be done correctly, a model must preserve the hierarchical relationships between object parts and their corresponding wholes. It is not sufficient that all the pieces of an object are present in an image, they must also be oriented correctly with respect to one another. Convolutional Neural Networks (CNNs) are limited in their ability to model the spatial hierarchy between objects. Even if sub-sampling operations (e.g., max-pooling) are removed, the representation of data in CNNs do not take into account the relationships between object parts and wholes.

Capsule Networks \cite{Sabour_2017}, \cite{Hinton_2018} learn such hierarchies by grouping feature map channels together to form a vector of features (i.e., a capsule) that captures the instantiation parameters of objects and by learning transformation matrices that encode viewpoint \textit{invariant} information. These networks generalize to novel viewpoints by incorporating the viewpoint changes directly into the activities of the capsules. The capsules can represent various properties of an object ranging from its size and shape to more subtle features such as texture and orientation. Since their introduction, CapsNets have produced state-of-the-art accuracies on datasets such as MNIST \cite{Zhao_2019} and smallNORB using a network with far fewer parameters compared with their CNN counterparts \cite{Sabour_2017}.

At their core, CapsNets \cite{Sabour_2017} make use of a dynamic routing algorithm to ensure that the output of a lower-level capsule is routed to the most appropriate higher-level capsule. This is done by multiplying the lower-level capsules with learned viewpoint invariant transformation matrices to produce prediction vectors. These matrices make sense of the spatial relationships between object features. The scalar product between the prediction vectors and each of the higher-level capsules governs the agreement between a part and a whole. Large values imply that a part has a high likelihood of belonging to a whole and vice versa. The combination of the scalar products and the transformation matrices ultimately decides which whole is most suited for each part. This ``routing-by-agreement'' is a more effective way of sending spatial information between layers in a network than the routing implemented by max-pooling, since the former maintains the exact spatial relationships between neurons in each layer of the network, regardless of the network depth.

Although CapsNets have shown promising results, there are some key limitations that make them unsuitable for real world use. One such issue is inference using CapsNets is significantly slower compared with CNNs. This is primarily due to the dynamic routing procedure which requires several iterations to produce the output vectors of a capsule layer. This limitation prevents deeper CapsNet architectures from being used in practice.

We present a method for speeding up inference for CapsNets, with potential applications for training. This is accomplished by making use of the accumulated information in the dynamically calculated routing coefficients computed using the training dataset. Analyses of intra-class routing coefficients show that they form a unique signature for each class. Intuitively, this is because parts of the same object should generally be similar and, more importantly, distinct from the parts of a different object. In practice, the network is trained to produce prediction vectors (using the lower-level capsules and learned transformation matrices) that closely correlate with the higher-level capsule associated with its own class. This observation allows creation of a set of \textit{master} routing coefficients using the individual routing coefficients associated with each training example. At inference time, the master routing coefficients are used instead of dynamically computing the routing coefficients for each input to the network. Our method for fast routing at inference effectively replaces the \boldmath{$r$} iterations in the routing procedure with a single matrix-multiply operation, allowing the network to be parallelized at run-time. On the MNIST dataset and its variants, fast inference decreases the test time accuracies by less than $0.5 \%$ and by approximately $5 \%$ for CIFAR10.

Section \ref{Section: Capsule Network Architecture} describes the three-layer network architecture that was used. Section \ref{Section: Comparison Between Dynamic and Fast Routing Procedures} compares the differences in the routing procedure at inference between the dynamic routing algorithm and the fast routing procedure. Section \ref{Section: Analyses of Dynamically Calculated Routing Coefficients} analyses the dynamic routing coefficients from the MNIST and CIFAR10 training datasets. In Section \ref{Section: Creation of Master Routing Coefficients}, we detail the procedure for creating a set of master routing coefficients and compare the master set of routing coefficients with the dynamically calculated ones in Section \ref{Section: Analyses of Master Routing Coefficients}. Section \ref{Section: Results} compares the test accuracies between the dynamic and fast routing procedures for the five datasets. Discussions on the use and applicability of master routing coefficients are given in Section \ref{Section: Discussion}. The general procedure for creating master routing coefficients is detailed in Appendix \ref{App. General Approach for Creating Master Routing Coefficients}.

\section{Capsule Network Architecture}
\label{Section: Capsule Network Architecture}
\begin{figure}[htp]
\centering
{\includegraphics[width = 3.5 in]{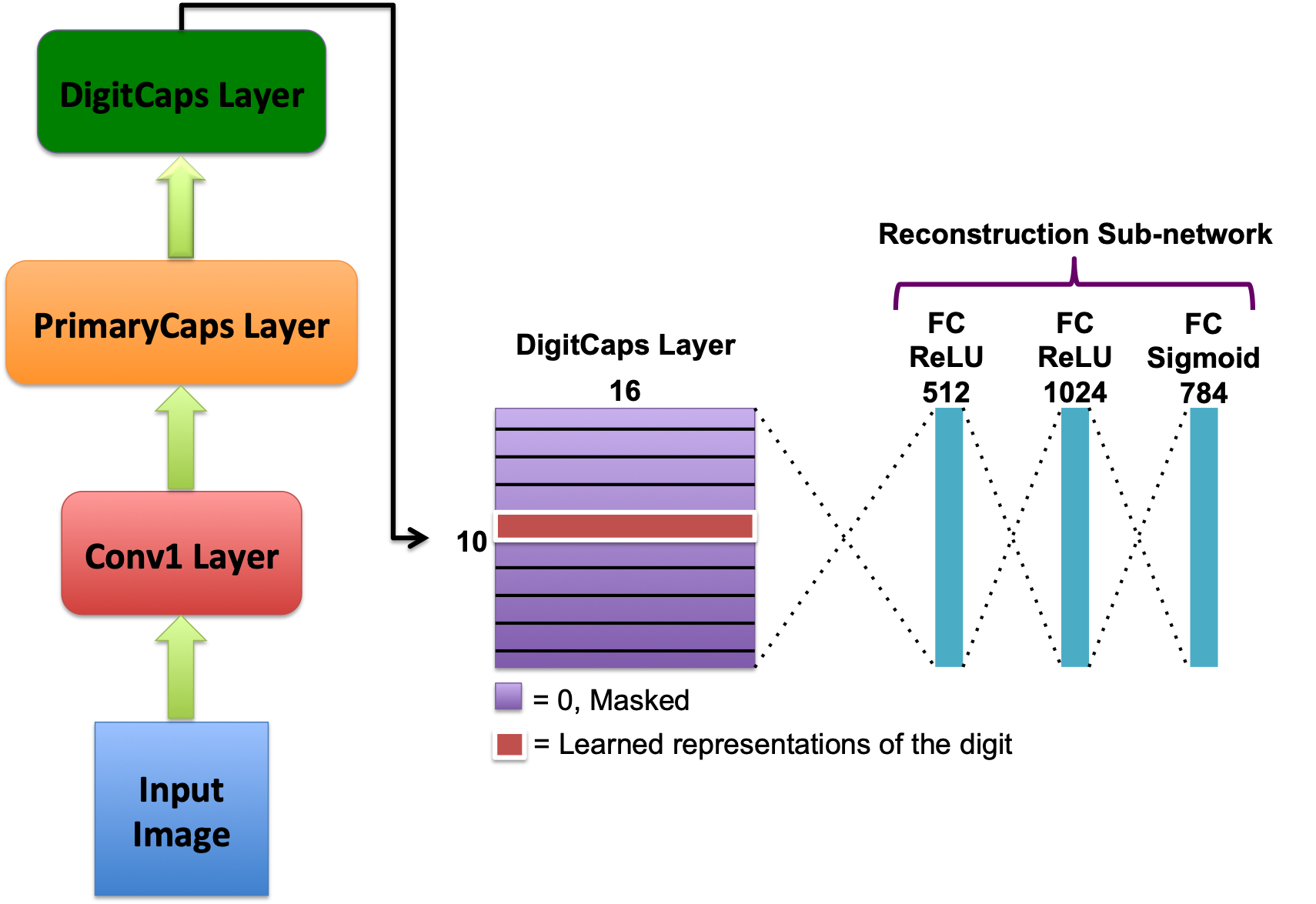}}
\caption{(Left) Three-layer CapsNet architecture adapted from Sabour et al. \cite{Sabour_2017}. The PrimaryCaps layer consists of $6 \times 6 \times 32 = 1152$ $8$-D vector capsules for the MNIST dataset and its variants ($1024$ $4$-D vector capsules for CIFAR10). The routing procedure produces the $10 \times 16$ DigitCaps layer, which is used to calculate the margin loss. The DigitCaps output is also passed to the reconstruction network where the non-ground-truth rows are masked with zeros. The network takes the masked $10 \times 16$ DigitCaps matrix as input and learns to reproduce the original image. Margin and reconstruction loss functions follow those from \cite{Sabour_2017}.}
\label{Fig: CapsNet Architecture}
\end{figure}

The CapsNet architecture shown in Fig. \ref{Fig: CapsNet Architecture} follows the three layer network from \cite{Sabour_2017}.  For the MNIST \cite{MNIST}, Background MNIST (bMNIST) and Rotated MNIST (rMNIST) \cite{R_and_BG_MNIST}, and Fashion MNIST (fMNIST) \cite{F_MNIST} datasets, the input to the network is a $28 \times 28$ grayscale image that is operated on by a convolutional layer to produce a $20 \times 20 \times 256$ feature map tensor (for CIFAR10, the $32 \times 32 \times 3$ image is randomly cropped so that its spatial dimensions are $24 \times 24$). The second convolutional layer outputs a $6 \times 6 \times 256$ feature map tensor. Each group of $8$ neurons ($4$ for CIFAR10) in this feature map tensor is then grouped channel-wise and forms a single lower-level capsule, \boldmath{$i$}, for a total of $6 \times 6 \times (256 \div 8) = 1152$ lower-level capsules ($1024$ for CIFAR10).

The lower-level capsules are fed to the routing layer, where the dynamic routing procedure converts these capsules to the $10 \times 16$ DigitCaps matrix, where $10$ is the number of higher-level capsules (also the number of object classes) and $16$ is the dimensionality of the capsules. Here, we use Max-Min normalization \cite{Zhao_2019} as opposed to Softmax to convert the raw logits into routing coefficients. Details on the dynamic routing procedure are given in Section \ref{Section: Comparison Between Dynamic and Fast Routing Procedures}.

Each row in the DigitCaps matrix represents the instantiation parameters of a single object class and the length of the row vector represents the probability of the existence of that class. During training, non-ground-truth rows in the DigitCaps matrix are set to zero and the matrix is passed to a reconstruction network composed of two fully-connected layers of dimensions $512$ and $1024$ with ReLU activations, and a final fully-connected layer of dimension $28 \times 28 \times 1 = 784$ ($24 \times 24 \times 3 = 1728$ for CIFAR10) with a sigmoid activation. During inference, the reconstruction network is not used. Instead, the row in the DigitCaps matrix with the largest L2-norm is taken as the predicted object class for the input.

Our implementation uses TensorFlow \cite{TensorFlow} with training conducted using the Adam optimizer \cite{Adam_Optimizer} with TensorFlow's default parameters and an exponentially decaying learning rate. Unless otherwise noted, the same network hyperparameters in \cite{Zhao_2019} were used here for training as well. Original code is adapted from \cite{Sabour_Code}.

\section{Comparison Between Dynamic and Fast Routing Procedures}
\label{Section: Comparison Between Dynamic and Fast Routing Procedures}
The dynamic routing procedure for CapsNets using Max-Min normalization is given by Max-Min Routing Procedure below. This algorithm is used during normal training and inference. The prediction vectors to the routing layer, \boldmath{$\hat{u}_{j|i}$}, are created by multiplying each of the lower-level capsules, \boldmath{$u_i$}, in PrimaryCaps by their respective transformation weight matrix, \boldmath$W_{ij}$. The higher-level capsules \boldmath{$s_j$} are computed as a sum over all the prediction vectors, weighted by the routing coefficients, \boldmath{$c_{ij}$}. The routing coefficients are initialized with a value of $1.0$ and can be viewed as independent probabilities representing the likelihood of a lower-level capsule being assigned to a higher-level capsule \cite{Zhao_2019}. For a given input to the routing layer, its routing coefficient matrix has shape $N_i \times N_j$, where $N_i$ and $N_j$ are the number of lower and higher-level capsules, respectively. The higher-level capsules, \boldmath{$s_j$}, are then squashed using a non-linear function so that the vector length of the resulting capsule, \boldmath{$v_j$}, is between $0$ and $1$. These operations are shown in Eq. \ref{Eqs. u_hat, s_J, and v_J}.

\begin{equation}
	\label{Eqs. u_hat, s_J, and v_J}
	\hat{u}_{j|i} = W_{ij}u_i, \quad s_j = \sum_{i} {c_{ij}\hat{u}_{j|i}}, \quad v_j = \frac{||s_j||^2}{1 + ||s_j||^2}\frac{s_j}{||s_j||}
\end{equation}

During the procedure, the update to the routing coefficients, \boldmath{$b_{ij}$}, is computed as the dot product between the prediction vectors and the current state of the higher-level capsules, \boldmath{$v_j$}. The update to the routing coefficients is then normalized via Max-Min normalization over the object classes as given by Eq. \ref{Eq: Max-Min Normalization}, where $p$/$q$ are the lower/upper bounds of the normalization. For the first iteration, the routing coefficients are initialized to $1.0$ outside of the main routing for-loop.

\begin{equation}
	\label{Eq: Max-Min Normalization}
	c_{ij} = p + \frac{b_{ij} - min(b_{ij})}{max(b_{ij}) - min(b_{ij})} * (q-p)
\end{equation}

\begin{table}[h!]
    \begin{tabular}{l}
    \hline
    \textbf{Max-Min Routing Procedure} \\
    \hline
    1: Input to Routing Procedure: ({$\bm{\hat{u}_{j|i}}$}, $r$, $l$) \\
    2: \quad for all capsules $i$ in layer $l$ and capsule $j$ in layer ($l$ + 1): $c_{ij}$ $\leftarrow$ 1.0 \\
    3: \quad \textbf{for} $r$ iterations: \\
    4: \quad \quad for all capsule $j$ in layer ($l$ + 1): $\bm{s_j}$ $\leftarrow$ $\sum_{i} c_{ij} \bm{\hat{u}_{j|i}}$ \\
    5: \quad \quad for all capsule $j$ in layer ($l$ + 1): $\bm{v_j}$ $\leftarrow$ Squash($\bm{s_j}$) \\
    6: \quad \quad for all capsule $i$ in layer $l$ and capsule $j$ in layer ($l$ + 1): $b_{ij} \leftarrow b_{ij} + \bm{\hat{u}_{j|i}} \cdot$ $\bm{v_j}$ \\
    7: \quad \quad for all capsule $i$ in layer $l$: $\bm{c_i}$ $\leftarrow$ Max-Min ($b_{ij}$) $\Longrightarrow$ $\mathrm{Given~in~Eq.}$ ~\ref{Eq: Max-Min Normalization} \\
       \quad \quad \textbf{return} $\bm{v_j}$ \\
    \hline
    \end{tabular}
    \label{Procedure: Max-Min Routing}
\end{table}

For fast inference, the routing coefficients no longer need to be dynamically calculated for each new input. Instead, the prediction vectors are simply multiplied with the precomputed master routing coefficient tensor ($C_{ij}$), summed, and squashed to form the higher-level capsules. Classification using the parent-level capsules, \boldmath{$v_j$}, is made in the usual way afterwards. Details on the master routing coefficients are given in the following sections.

\begin{table}[h!]
    \begin{tabular}{l}
    \hline
    \textbf{Fast Routing Procedure for Inference} \\
    \hline
    1: Input to Fast Routing Procedure: ($C_{ij}$, \boldmath$\hat{u}_{j|i}$, $l$) \\
    2: \quad for all capsule $j$ in layer ($l$ + 1): $s_j$ $\leftarrow$ $\sum_{i} C_{ij}\boldmath\hat{u}_{j|i}$ \\
    3: \quad for all capsule $j$ in layer ($l$ + 1): $v_j$ $\leftarrow$ Squash($s_j$) \\
       \quad \quad \textbf{return} $v_j$ \\
    \hline
    \end{tabular}
    \label{Procedure: Fast Routing for Inference}
\end{table}

\section{Analyses of Dynamically Calculated Routing Coefficients}
\label{Section: Analyses of Dynamically Calculated Routing Coefficients}
For each input to the routing layer, an $N_i \times N_j$ routing coefficient matrix is initialized with a value of $1.0$ and then iteratively updated in the routing procedure. By design, the $i^{th}$ column of routing coefficients is responsible for linking the lower and higher-level capsules for the $i^{th}$ object class. Intuitively, one would expect intra-class objects to have similarly activated routing coefficients compared with those from inter-class objects since the routing coefficients are simply the agreement between prediction vectors and higher-level capsules. This can be shown quantitatively by computing the correlations between the dynamically calculated routing coefficients for the different object classes.

We choose to compute the correlations between only the ground-truth (GT) columns in the routing coefficient matrices rather than between entire routing coefficient matrices, since the GT columns are used for the correct classification of each image. In other words, for a network that has been properly trained, the routing coefficients in the GT columns are the ones that provide the unique signatures for each object class. The routing coefficients in the other columns generally do not evolve to form unique signatures in the same manner as the GT routing coefficients. This can be observed in the tuning curves associated with the higher-level capsules (c.f. Section \ref{Section: Analyses of Master Routing Coefficients}).

\begin{figure}[h]
\centering
{\includegraphics[width = 4.5 in]{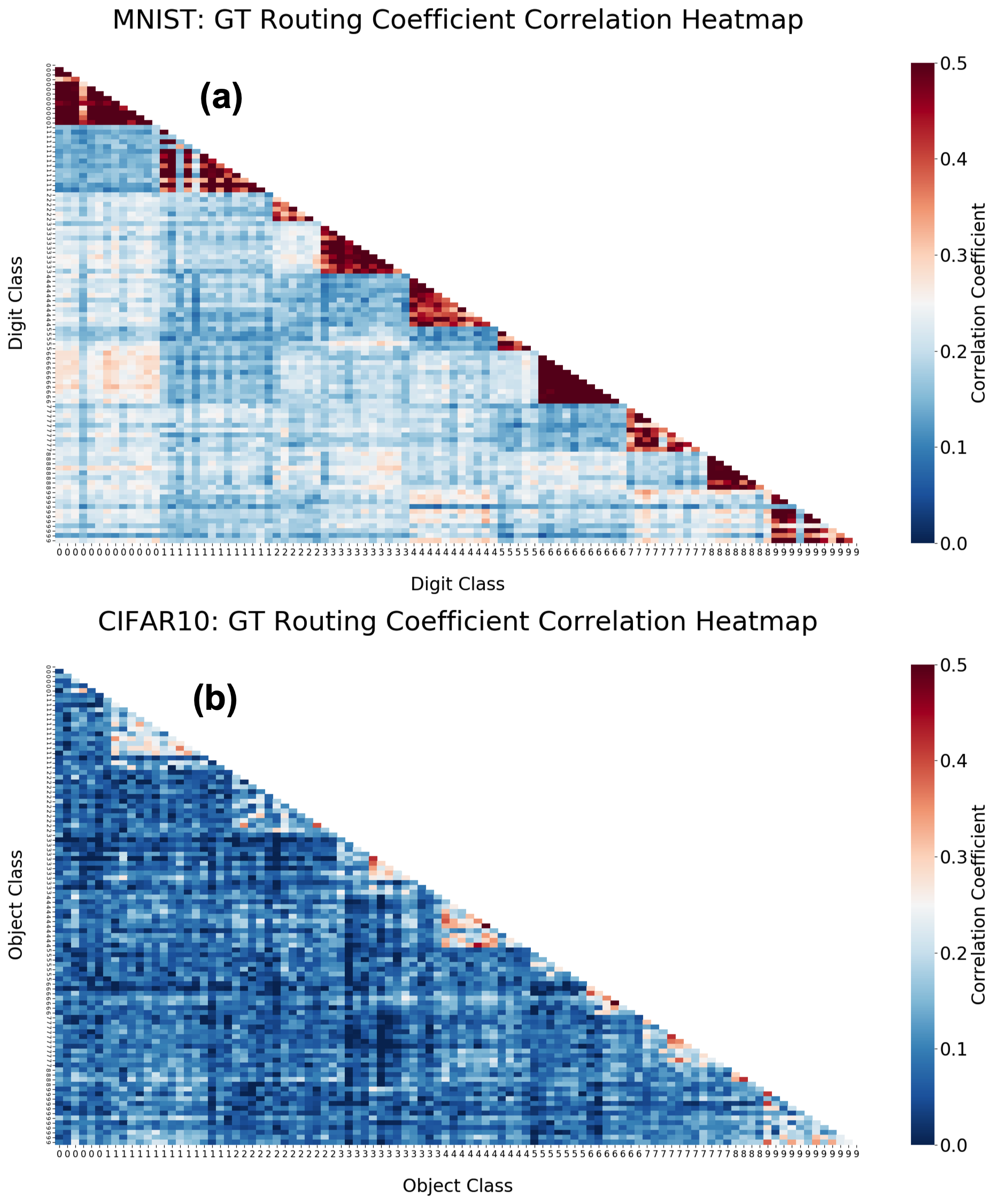}}
\caption{(a) Heatmap showing the correlations between the GT columns for the first $100$ images in the MNIST training set. The heatmap is created by calculating the correlation coefficient between GT columns in the $1152 \times 10$ routing coefficient matrix associated with each image. (b) Heatmap showing the correlations between the GT columns for the first $100$ images in the CIFAR10 training set. The heatmap is created by calculating the correlation coefficient between GT columns in the $1024 \times 10$ routing coefficient matrix associated with each image. Only half of the computed values are shown for each heatmap since the correlations are symmetric.}
\label{Fig: GT Routing Coefficient CH}
\end{figure}

Figure \ref{Fig: GT Routing Coefficient CH} (a) shows the correlation heatmap between the GT columns of the routing coefficient matrices for the first $100$ images in the MNIST training dataset. For each image, the GT column in its routing coefficient matrix refers to the column that corresponds to its object class. For example, an image of the digit $5$, the GT column in that image's routing coefficient matrix is the sixth column (due to zero-indexing). Likewise, Fig. \ref{Fig: GT Routing Coefficient CH} (b) shows the correlation heatmap for the first $100$ images in the CIFAR10 training dataset calculated in the same way. For MNIST, high correlation is observed between routing coefficients for intra-class objects. For a more complicated dataset such as CIFAR10, the distinction between inter-class routing coefficients is not as clear since the same part can often be associated with more than one object class in the dataset. For example, trucks and automobiles will have often multiple similar parts (e.g., wheels, headlights, body frame, etc.) and airplanes and ships will often appear against the same color background.

\begin{figure}[h]
\centering
{\includegraphics[width = 4.5 in]{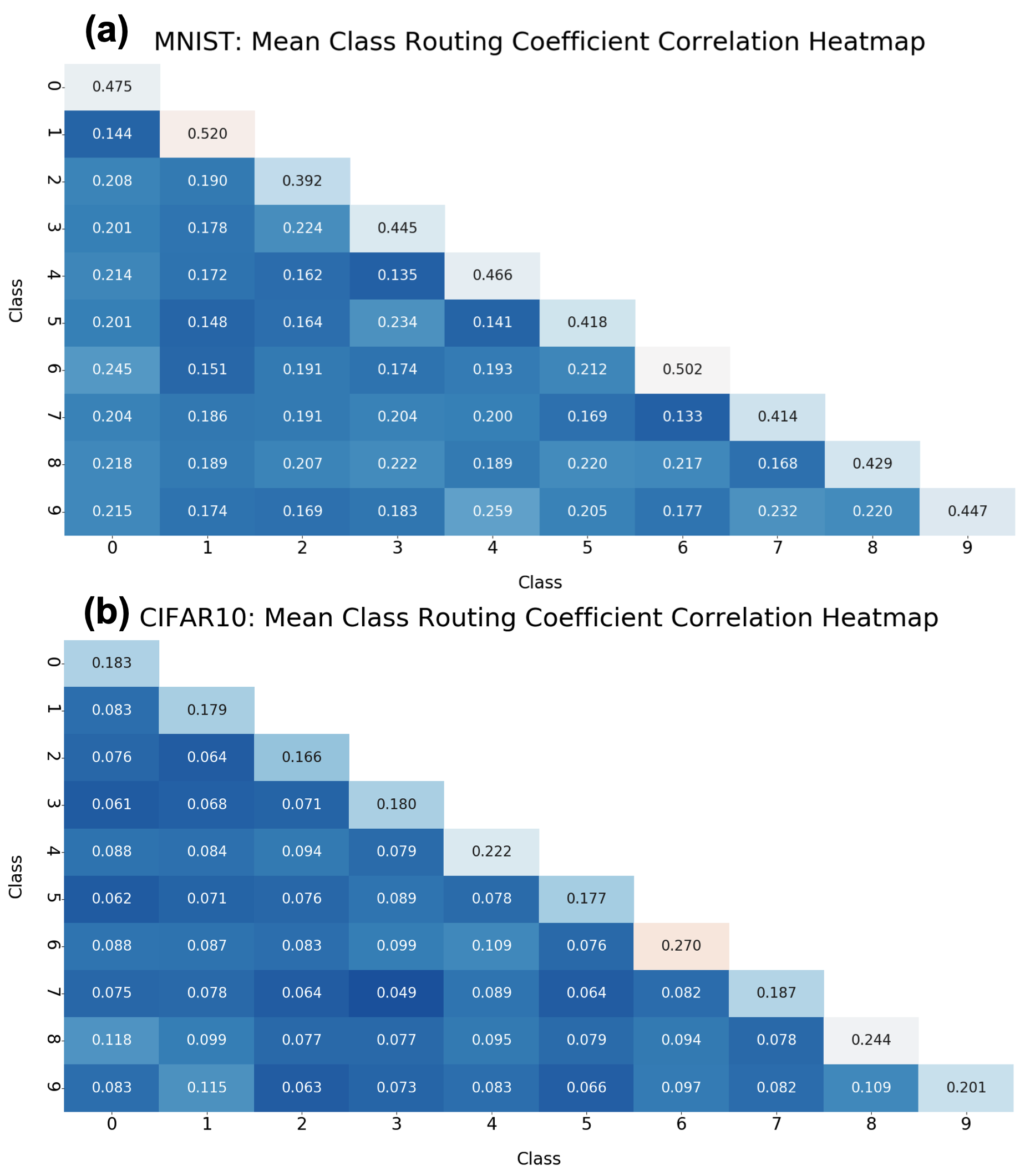}}
\caption{(a) Heatmap showing the average class correlations for \textit{all} images in the MNIST training set. For example, the value in element $(0, 0)$ is the mean correlation between the GT columns in the routing coefficient matrix for all objects in class $0$ and the value in element $(0, 1)$ is the mean correlation between the GT columns in the routing coefficient matrix for all objects in class $0$ and $1$, etc. (b) Heatmap showing the average class correlations for \textit{all} images in the CIFAR10 training set. Only half of the computed values are shown for each heatmap since the correlations are symmetric.}
\label{Fig: Mean Class Routing Coefficient CH}
\end{figure}

Figures \ref{Fig: Mean Class Routing Coefficient CH} (a) and (b) show the correlation heatmaps between each of the object classes for the MNIST and CIFAR10 datasets, respectively, averaged over \textit{all} of their training images. For example, element $(0, 0)$ in the heatmaps is the mean correlation between the GT columns of all objects in class $0$ for the dataset and element $(0, 1)$ is the mean correlation between the GT columns of all objects in classes $0$ and $1$, and so on. The same trends exist when the correlation is computed across all training images as those in Fig. \ref{Fig: GT Routing Coefficient CH} for the case of $100$ images for each dataset. In particular, it is interesting to note the second highest correlations for each object class. For the MNIST dataset, the second highest correlation with the digit $0$ is the digit $6$ and the digit $9$ has high correlations with the digits $4$ and $7$. These digit classes are often most similar to one another for the MNIST dataset (c.f. \cite{Zhao_2019} for examples). For CIFAR10, high inter-class correlations exist between airplanes (class $0$) and ships (class $8$) and automobiles (class $1$) and trucks (class $9$). These classes often present the most challenging examples for classification. For MNIST, intra-class correlations are significantly higher compared with inter-class correlations and thus, the network is able to properly distinguish between the digit classes. For CIFAR10, intra-class correlations are lower and this leads to difficulties in classifying new images.

\section{Creation of Master Routing Coefficients}
\label{Section: Creation of Master Routing Coefficients}
\begin{figure}[h]
\centering
{\includegraphics[width = 4.5 in]{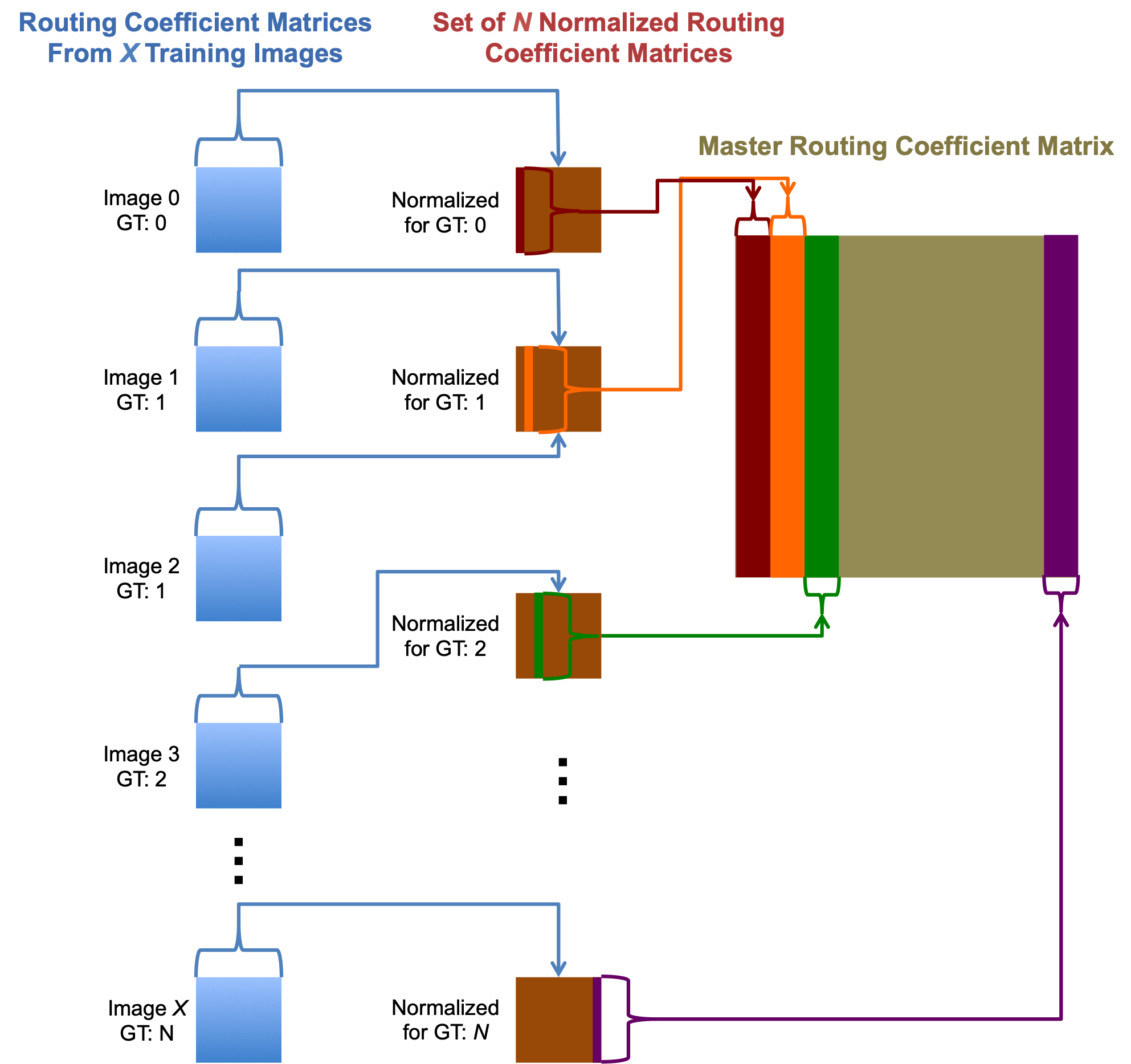}}
\caption{Procedure for creating a single master routing coefficient  matrix for fast inference. After a network has been properly trained, the routing coefficient matrix associated with each training image is extracted and used to form a single master routing coefficient matrix via three steps: 1) summation, 2) normalization, and 3) dimension reduction. Details are given in Section \ref{Section: Creation of Master Routing Coefficients}.}
\label{Fig: Creation of Master Coefficients}
\end{figure}

Since the intra-class routing coefficients from the training dataset form a unique signature for each object class, they can be used to create a set of master routing coefficients that can generalize well to new examples. There are several ways in which a set of master routing coefficients can be created. We detail the procedure shown in Fig. \ref{Fig: Creation of Master Coefficients}, which was used to generate the master routing coefficients used for fast inference. A general approach for creating master routing coefficients is given in Appendix \ref{App. General Approach for Creating Master Routing Coefficients}.

During training, the routing coefficient matrix is initialized to $1.0$ for each input to the routing layer and then iteratively updated to reflect the agreement between the input prediction vectors to the routing layer and the final output vectors of the routing layer. At the start of training, the routing coefficients can change for the same input since the prediction vectors are being updated by the (trainable) network weights, \boldmath$W_{ij}$. However, once training has converged, the routing coefficients naturally converge since they are computed in a bootstrap manner; i.e., the update to the routing coefficients are calculated using just the prediction vectors and an initial set of routing coefficients. After training is completed, the routing coefficient matrix associated with each training image can be extracted.

For training images, fast inference can be conducted on the \textit{training} dataset by simply using the individual routing coefficient matrix associated with each image. For new images, however, there is no apriori method of determining which routing coefficient matrix to use for each image since the class label is unknown (if such a method existed, then the classification problem is essentially solved without the need for the network to even make the prediction). As a result, fast inference on new images must rely on the use of a \textit{single} routing coefficient matrix. This matrix has the same shape as the individual routing coefficient matrices and each column is assigned to a known object class.

To create the master routing coefficient matrix, we accumulate the information contained in the individual routing coefficient matrices from the training dataset. This process involves three main steps: 1) summation, 2) normalization, and 3) dimension reduction. First, we train the CapsNet to convergence in the usual manner, run inference on the training images, and save the routing coefficients associated with the training images at the last step of the routing procedure (for MNIST, this results in $60,000$ $1152 \times 10$ matrices). Then, we initialize $N$ matrices as containers to hold the accumulated class-specific routing coefficients for each of the $N$ object classes. For each training example, its individual routing coefficient matrix is summed in the appropriate container matrix for its class. After all training images have been processed, the set of $N$ container matrices holds the sum of all routing coefficients at the last routing iteration, one for each class.

Each container matrix is then normalized by their respective class frequency, followed by a row-wise Max-Min normalization. At this point, each container matrix can be viewed as a master routing coefficient matrix for that class. However, the network expects a single routing coefficient matrix and, without additional information, there is no straightforward method to point to the correct container matrix when the network is presented with a new example.

Thus, in order to present a single routing coefficient matrix to the network at inference, the set of $N$ container matrices must be reduced to a single matrix. This is done by transferring only the GT column from each of the $N$ container matrices to its corresponding column in the master routing coefficient matrix. In other words, only the first column from the first container matrix (which holds the accumulated routing coefficients for the first object class) is transferred to the first column of the master routing coefficient matrix, and so on for the columns in the other container matrices. The end result of the dimension reduction is a single routing coefficient matrix that can be used for new examples during inference.

\section{Analyses of Master Routing Coefficients}
\label{Section: Analyses of Master Routing Coefficients}
In order for the master routing coefficient matrix to generalize well to new images, each column of routing coefficients in the matrix should only correlate highly with its own object class. In other words, the first column of routing coefficients in the master routing coefficient matrix should correlate highly with routing coefficients in the first columns of the individual routing coefficient matrices for images associated with the first object class. The second column of routing coefficients in the master routing coefficient matrix should correlate highly with routing coefficients in the second columns of the individual routing coefficient matrices for images associated with the second object class, and so on. If the first column in the master routing coefficient matrix correlated highly with the fifth column in an individual routing coefficient matrix associated with an image from the fifth object class, this would imply that the first column of master routing coefficients does \textit{not} form an unique signature for the first object class---it can likely classify an image that belongs to the first object class as that of the fifth object class (and vice versa).

\begin{figure}[h]
\centering
{\includegraphics[width = 4.5 in]{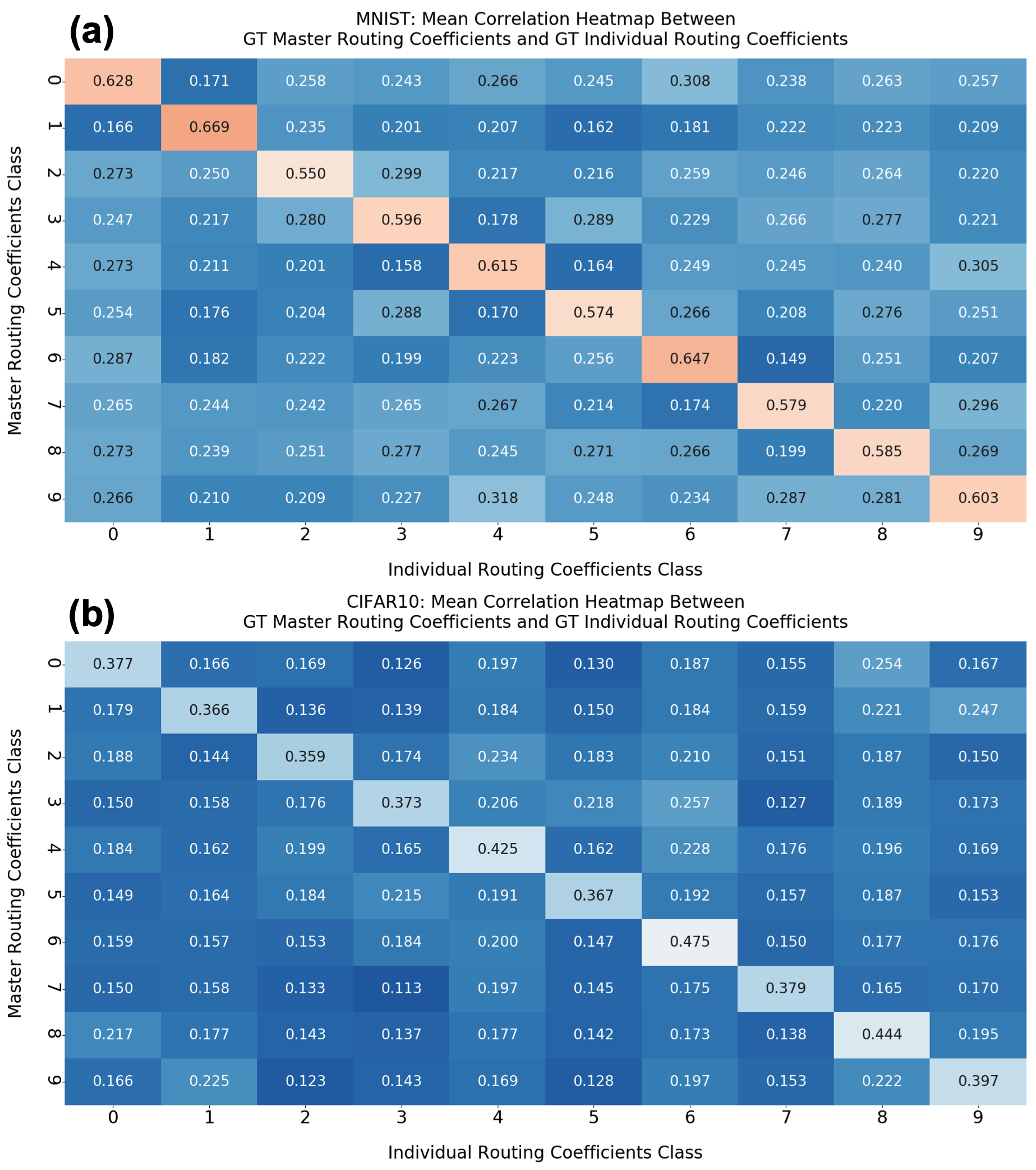}}
\caption{(a) Correlation heatmap between the columns in the master routing coefficients matrix and the GT columns from the individual routing coefficient matrices associated with the MNIST training images. This shows that creating the master routing coefficients as outlined in Section \ref{Section: Creation of Master Routing Coefficients} results in a set of routing coefficients that correlates highly within its own class. (b) Correlation heatmap between the columns in the master routing coefficients matrix and the GT columns from the individual routing coefficient matrices associated with the CIFAR10 training images. For CIFAR10, high correlations can also exist between \textit{inter}-class routing coefficients. For example, the correlation between the routing coefficients in column $1$ (associated with the object class ``car'') of the master routing coefficient matrix and GT columns $9$ (associated with the object class ``truck'') from individual routing coefficient matrices is almost as high as the \textit{intra}-class correlations for classes $1$ and $9$, individually.}
\label{Fig: Mean CH Between GT Master and GT Individual Routing Coefficients}
\end{figure}

To quantify the degree to which each column of the master routing coefficient matrix is representative of its object class, we compute the correlation between each column in the master routing coefficient matrix and the GT columns in each individual routing coefficient matrix. This is shown in Figs. \ref{Fig: Mean CH Between GT Master and GT Individual Routing Coefficients} (a) and (b) for the MNIST and CIFAR10 datasets, respectively. These correlations are between a single column of routing coefficients from the \textit{master} routing coefficient matrix and the GT column of routing coefficients from the \textit{individual} routing coefficient matrix associated with each training image for the dataset. Thus, the correlations are not symmetric (i.e., the correlation for element $(0, 1)$ in the heatmap is not the same as the correlation for element $(1, 0)$).

From Figs. \ref{Fig: Mean CH Between GT Master and GT Individual Routing Coefficients} (a) and (b), we see that each column of routing coefficients in the master routing coefficient matrix has higher intra-class correlations than inter-class correlations with the individual GT routing coefficients from each image in the training dataset. This is effectively why the master routing coefficients are able to generalize well to new images during inference---they are able to uniquely route the lower-level capsules to the correct higher-level capsules for new images by using the accumulated information from the training data. This relationship between master and individual routing coefficients is stronger for simpler datasets such as MNIST and its variants than for a more complicated dataset such as CIFAR10. For CIFAR10, intra-class correlations are still higher compared with inter-class correlations; however, as mentioned above, inter-class correlations can also be high for objects belonging to similar classes.

\begin{figure}[h]
\centering
{\includegraphics[width = 5.5 in]{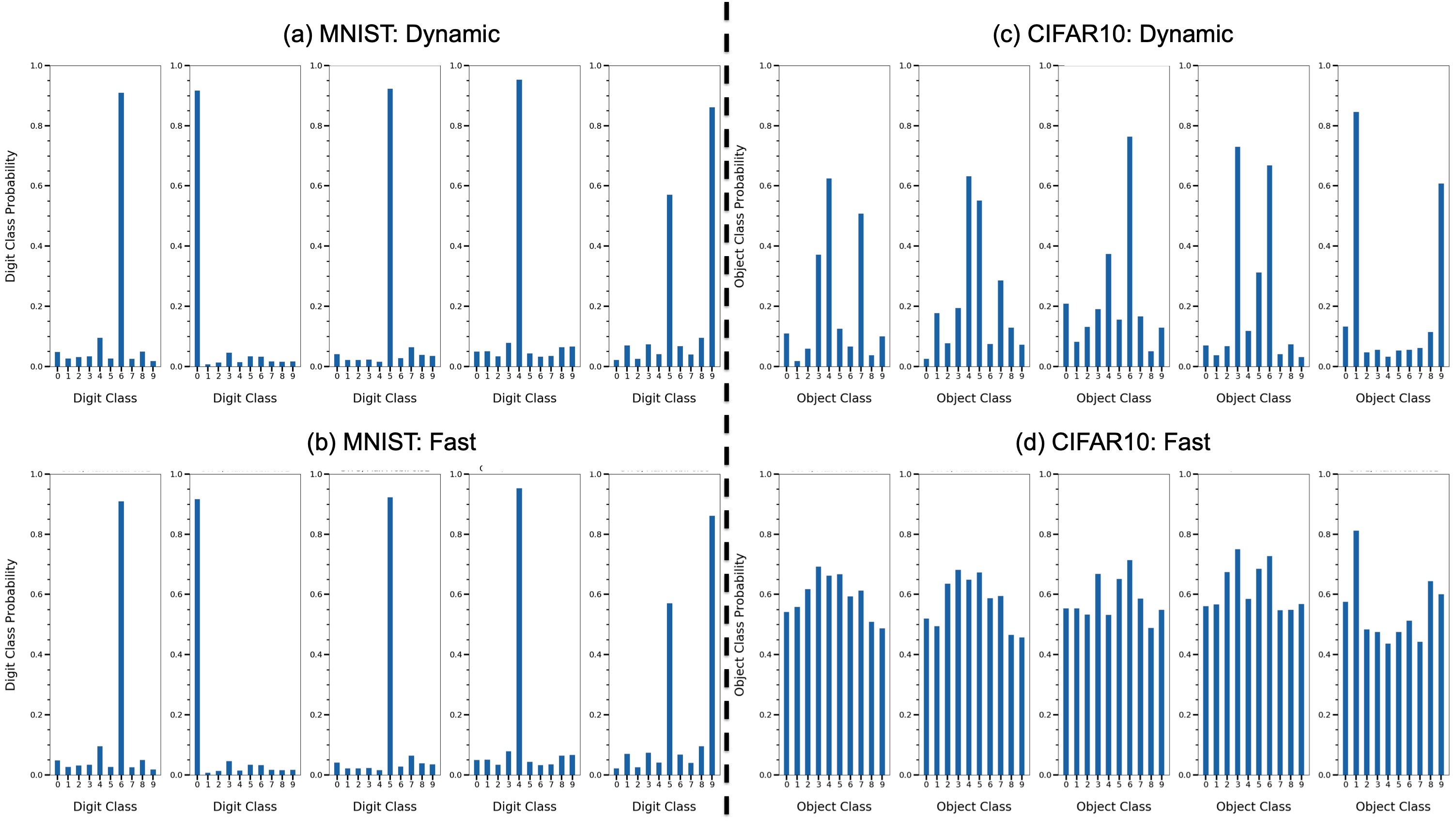}}
\caption{Output class probabilities for the same set of test images from the MNIST and CIFAR10 datasets. The class probabilities are obtained from networks that use dynamically calculated routing coefficients ((a) and (c)) and the master routing coefficients ((b) and (d)). For MNIST, the master routing coefficients produce nearly identical class probabilities compared with the use of dynamically calculated routing coefficients. For CIFAR10, the use of master routing coefficients correctly classify three out of the five test image examples shown and there is strong inter-class competition.}
\label{Fig: DigitCaps Routing Examples}
\end{figure}

The effectiveness of the master routing coefficients can also be examined by looking at the output capsules, \boldmath{$v_j$}, from the DigitCaps layer. Figures \ref{Fig: DigitCaps Routing Examples} (a) and (b) compare the digit class probabilities for the same set of test images from the MNIST dataset between the dynamic and fast inference routing procedures. Likewise, Figs. \ref{Fig: DigitCaps Routing Examples} (c) and (d) compare the probabilities for the same set of test images from the CIFAR10 dataset. For MNIST, the master routing coefficients produce similarly peaked values for the digit class probabilities compared with the use of dynamically calculated routing coefficients---each column in the master routing coefficient matrix is a unique signature for that digit class. On the other hand, the digit class probability comparison for CIFAR10 is noticeably different. Although the master routing coefficients are able to correctly classify three out of the five test examples shown, the classification is not particularly robust compared with the use of dynamically calculated routing coefficients (the dynamically calculated routing coefficients correctly classifies four out of the five test image examples shown, and have lower probabilities for the non-ground-truth classes).

\begin{figure}[h]
\centering
{\includegraphics[width = 5.5 in]{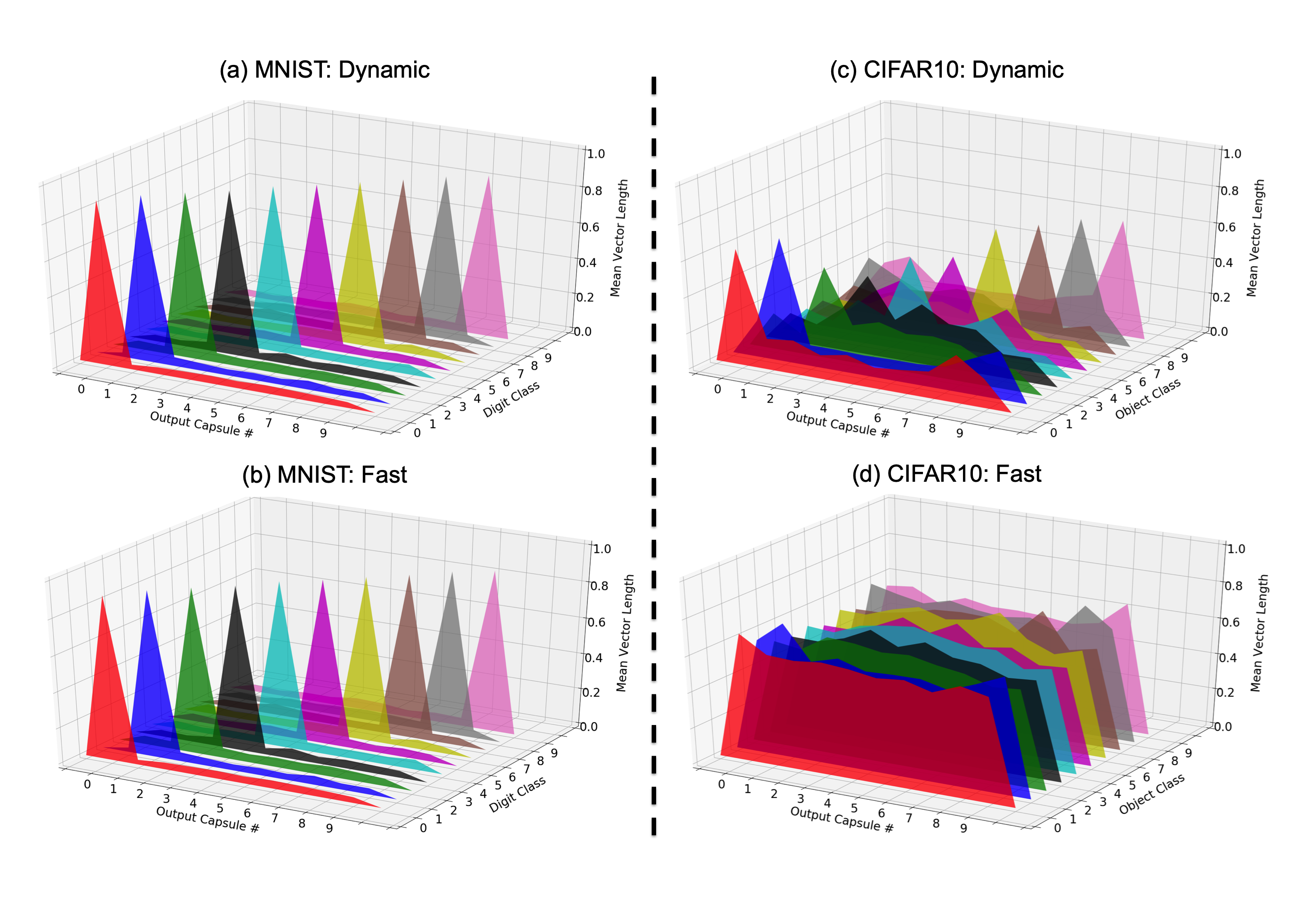}}
\caption{Class-averaged output probabilities for the MNIST and CIFAR10 test datasets obtained with the use of dynamically calculated routing coefficients ((a) and (c)) and the master routing coefficients ((b) and (d)). The master routing coefficients for the MNIST dataset produce tuning curves similar to those obtained using dynamically calculated routing coefficients. For CIFAR10, the tuning curve for each object class obtained using dynamically calculated routing coefficients exhibit multiple peaks around non-ground-truth classes for each object class. In addition, multiple high-valued peaks are observed for the tuning curves resulting from the use of master routing coefficients.}
\label{Fig: Mean Vector Length Per Class}
\end{figure}

The digit class probabilities can also be class-averaged over the test dataset as shown in Fig. \ref{Fig: Mean Vector Length Per Class}. These ``tuning curves'' show how well the network is able to distinguish between the different object classes in the dataset. For MNIST, the tuning curves resulting from the fast inference procedure is similar to those from the dynamic routing procedure, suggesting that the master routing coefficients provides an accurate reflection of the dynamically calculated routing coefficients. The tuning curves for CIFAR10 from the dynamic routing procedure show that, for certain object classes, the discriminability is not as robust compared with MNIST---multiple peaks exist for each object class. For fast inference, each tuning curve is still peaked at the correct object class; however, they are also highly peaked around the other classes as well.

\section{Results}
\label{Section: Results}
Table \ref{Table: Routing Method Accuracies} provides the test accuracies for five datasets for the two different routing methods at inference. For fast routing, the master routing coefficients are created using the steps detailed in Section \ref{Section: Creation of Master Routing Coefficients} and the procedure is given by Fast Routing Procedure for Inference in Section \ref{Section: Comparison Between Dynamic and Fast Routing Procedures}. With Max-Min normalization, fast routing decreases the test accuracy by approximately $0.5 \%$ for the MNIST dataset and its variants and by $5 \%$ for CIFAR10.

The master routing coefficients can also be created from a network trained using Softmax as the normalization in the routing layer as opposed to Max-Min. In this approach, the routing procedure is exactly the same as in \cite{Sabour_2017} and the master routing coefficients are created in the same manner as that detailed in Section \ref{Section: Creation of Master Routing Coefficients} except the Softmax function is used to normalize the rows in the container matrices instead of Max-Min. For networks trained with Softmax in the routing layer, the majority of the routing coefficients are grouped around their initial value of $0.1$ after three routing iterations \cite{Zhao_2019}. Since the container matrices for each class are created via summation, class frequency averaging, and normalizing via Softmax, the resulting master routing coefficient matrix contains values close to $0.1$ (for a dataset with $10$ classes). This amounts to a uniform distribution for the master routing coefficients and results in a significantly reduced test time performance across all five datasets.

\begin{table}[h]
    \centering
    \caption{Mean of the maximum test accuracies and their standard deviations on five datasets for the different routing methods at inference. Five training sessions were conducted for each dataset. The conditions under which the datasets were trained are the same as \cite{Zhao_2019}. The master routing coefficients used for fast inference is created using the procedure outlined in Section \ref{Section: Creation of Master Routing Coefficients}.}
    \begin{tabular}{cccccc}
    \hline
    \textbf{Routing Method} & \textbf{MNIST {[}\%{]}} & \textbf{bMNIST {[}\%{]}} & \textbf{fMNIST {[}\%{]}} & \textbf{rMNIST {[}\%{]}} & \textbf{CIFAR10 {[}\%{]}} \\
    \hline
    Dynamic, Max-Min & 99.55 $\pm$ 0.02 & 93.09 $\pm$ 0.04 & 92.07 $\pm$ 0.12 & 95.42 $\pm$ 0.03 & 75.92 $\pm$ 0.27 \\
    Fast, Max-Min & 99.43 $\pm$ 0.08 & 92.93 $\pm$ 0.10 & 91.52 $\pm$ 0.20 & 95.04 $\pm$ 0.04 & 70.33 $\pm$ 0.36 \\
    Dynamic, Softmax & 99.28 $\pm$ 0.07 & 89.08 $\pm$ 0.21 & 90.52 $\pm$ 0.16 & 93.72 $\pm$ 0.09 & 73.65 $\pm$ 0.10 \\
    Fast, Softmax & 98.92 $\pm$ 0.30 & 84.34 $\pm$ 4.37 & 80.16 $\pm$ 7.35 & 84.13 $\pm$ 4.28 & 47.11 $\pm$ 8.70 \\
    \hline
    \end{tabular}
    \label{Table: Routing Method Accuracies}
\end{table}

\section{Discussion}
\label{Section: Discussion}
In CapsNets, routing coefficients form the link between capsules in adjacent network layers. Throughout the dynamic routing procedure, the updates to the routing coefficients come from the agreement between the prediction vectors and the higher-level capsules calculated using the prediction vectors. Since the prediction vectors are computed by a convolutional layer and are learned by the network, they capture the lower-level features for that object class. Using this information, we create a set of master routing coefficients from the training data that generalize well to test images.

If the network is properly trained (i.e., it is able to adequately distinguish between each object class) then the prediction vectors are unique and, as a result, the routing coefficients have high intra-class correlations. This is the case for MNIST and its variants. For a more complex dataset such as CIFAR10, the network does not learn sufficiently different prediction vectors for each object class during training. This is evident by comparing the tuning curves in Fig. \ref{Fig: Mean Vector Length Per Class} between MNIST and CIFAR10 for the case of \textit{dynamic} routing. As a result, the master routing coefficients created for CIFAR10 do not perform as well.

A better process for creating a master routing coefficient matrix is also possible. In this paper, the approach taken to create the master routing coefficients uses \textit{all} of the training data. For networks that can sufficiently distinguish between each object class, using all of the data makes sense. For CIFAR10, not all individual routing coefficient matrices are equally useful since some have high inter-class correlations. Therefore, a method of selecting the routing coefficients that have high intra-class and low inter-class correlations can be helpful. This can be done in several ways. For example, clustering analyses can be used to group routing coefficients that have high intra-class properties and remove training examples that result in ``outlier'' routing coefficients for each class. A similarity measure (e.g., correlation, dot product, etc.) can also be used to exclude outliers. Filtering methods will be taken up in future work.

Given that a single set of routing coefficients can be used for inference, a question to ask is whether or not it can also be useful for training. At least two approaches can be implemented for training using the master routing coefficients. First, a CapsNet can be retrained using the single master routing coefficient matrix to see if the network can learn to recognize the object classes better when it is able to use the cumulative knowledge (contained in the master routing coefficients) from all training data at once. Second, training can be expedited by initially training a CapsNet using dynamic routing on a carefully selected subset (or all) of the training data for a few epochs. Afterwards, the master routing coefficient matrix can be created from this training session and used to retrain the network (using all of the training data) to convergence. Both of these approaches will be taken up in future work.

\section{Summary}
\label{Section: Summary}
Capsule Networks have the ability to learn part-whole relationships for objects and can potentially generalize to novel viewpoints better than conventional CNNs. In contrast to CNNs, they can maintain the exact spatial relationships between the different components of an object by discarding sub-sampling layers in the network. State-of-the-art performance has already been demonstrated for the MNIST dataset by \cite{Zhao_2019}. However, CapsNets are slow to train and run in real-time due to the dynamic routing algorithm. In addition, state-of-the-art performance on more complicated datasets still present some challenges, possibly due to the prohibitively high cost of constructing deeper CapsNet models. In this work, we focused on methods to improve the speed of CapsNets for inference while still maintaining the accuracy obtained using the dynamic routing algorithm. To this end, we have implemented a method that allows for fast inference while keeping the test accuracy comparable to the dynamic implementation.

\appendix
\section{General Approach for Creating Master Routing Coefficients}
\label{App. General Approach for Creating Master Routing Coefficients}
Here we outline a general procedure for creating master routing coefficients. 

\begin{enumerate}
    \item Train a CapsNet using the training dataset and the dynamic routing procedure with Max-Min normalization (i.e., allow the dynamic calculation of routing coefficients for each input to the routing layer).
    
    \item Evaluate the trained network on the images in the training dataset and save the routing coefficients, \boldmath{$c_{ij}$}, for each image. That is, for each image, there is a set of routing coefficients that are dynamically calculated in the routing algorithm over \boldmath{$r$} iterations. For the routing procedure given by Max-Min Routing Procedure, the routing coefficients for a single image are contained in the tensor \boldmath{$c_{ij}$}, with dimensions (\boldmath{$r$}, $1152$, $10$) (dimensions are for images from the MNIST dataset). For a batch of images, \boldmath{$c_{ij}$} is a tensor with dimension (\boldmath{$N$}, \boldmath{$r$}, $1152$, $10$), where \boldmath{$N$} is the number of images in the batch. There is a choice of which routing iteration to use for the extraction of the routing coefficients from individual images. The routing coefficients from the first iteration corresponds to a uniform distribution (i.e., all routing coefficients are initialized to the same constant value). We find that the routing coefficients at the last iteration are the ones most suitable for use in constructing the master routing coefficients.
    
    \item After the routing coefficients for the training dataset are saved, a filtering operation can be done to differentiate between ``good'' and ``bad'' routing coefficients. Good routing coefficients are those that result in a minimal decrease in the model's accuracy on new images when used to create the master routing coefficients (vice versa for bad routing coefficients). Three approaches that can be taken to filter routing coefficients are:
    
    \begin{enumerate}
        \item \textbf{No filtering}. In this case, all routing coefficients are used to construct the master routing coefficients. This is the approach taken for the procedure detailed in Section \ref{Section: Creation of Master Routing Coefficients}.
        
        \item \textbf{Filtering based on clustering algorithms}. In this approach, a clustering algorithm (e.g., Expectation-Maximization or K-Means) is used to group intra-class routing coefficients with similar properties. After the grouping is performed, a selection is made to filter away outlier routing coefficients from each object class.
        
        \item \textbf{Filtering based on similarity measures}. A similarity measure (e.g. Pearson correlation coefficient, dot product, etc.) is used to select routing coefficients for each object class that have high intra-class similarities. For example, the routing coefficients associated with a person should be most similar with the routing coefficients in the person class and dissimilar from the routing coefficients associated with the car class. The routing coefficients that are most similar with respect to its own class are kept.
    \end{enumerate}
    
    \item After the individual routing coefficients have been filtered, a set of master routing coefficients can be constructed using the filtered set of individual routing coefficients. In this step, the goal is to create a single set of master routing coefficients that generalizes well to new images. One approach is to accumulate just the GT columns from the individual routing coefficient matrices into a single container matrix, then normalize column-wise by class frequency and row-wise by Max-Min to form the master routing coefficient matrix. This method can be done but results in poor performance when tested on new images.
    
    \medskip
    
    A key observation to make here is that \textit{each} of the higher-level capsules competes for the assignment of a lower-level capsule as a result of the Max-Min normalization (with Max-Min, the assignments can be viewed as independent probabilities). As a result, the information content in non-ground-truth columns are just as important as those in the GT columns. Thus, when constructing a single set of master routing coefficients, it is important to accumulate the routing coefficients from the entire individual routing coefficient matrix instead of just the GT columns. This is the approach described in Section \ref{Section: Creation of Master Routing Coefficients}.
    
    \item Following this approach, the individual routing coefficient matrices are summed in their respective class container matrix. After all the routing coefficient matrices have been summed, the container matrices are normalized by their respective class frequencies, followed by a row-wise Max-Min normalization. Then, the GT columns from each of the container matrices are transferred to their respective columns in the master routing coefficient matrix.

    \medskip

    Finally, the master routing coefficient matrix is replicated \boldmath{$N$} times to conform to the number of images per batch used in the network so that the dimensions of the master routing coefficient tensor presented to the network for inference is (\boldmath{$N$}, $1152$, $10$).
\end{enumerate}

\end{document}